\documentclass[runningheads]{llncs}
\usepackage{graphicx}
\usepackage{epsfig}
\usepackage{graphicx}
\usepackage{amsmath}
\usepackage{amssymb}
\usepackage{multirow}
\usepackage{tabu}
\usepackage{booktabs}
\usepackage{bbm}
\usepackage{diagbox}
\usepackage[english]{babel}
\usepackage{comment}

\usepackage{xcolor}
\usepackage{subfigure}
\usepackage{wrapfig}
\usepackage{cite}
\usepackage{diagbox}
\usepackage{array}
\usepackage{amstext}
\usepackage{makecell}
\usepackage{caption}
\usepackage{tablefootnote}
\usepackage{babel}
\usepackage{amsmath}
\usepackage{chngcntr}
\usepackage{placeins}
\usepackage{booktabs}       
\usepackage{amsfonts}       
\usepackage{nicefrac}  

\begin{document}
\title{Evaluating and Boosting Uncertainty Quantification in Classification}
\titlerunning{Evaluating and Boosting Uncertainty Quantification}
\author{Xiaoyang Huang\thanks{These authors have contributed equally: X. Huang and J. Yang} \and
Jiancheng Yang\inst{\star} \and
Linguo Li  \and
Haoran Deng \and
Bingbing Ni\thanks{Corresponding author: Bingbing Ni.} \and Yi Xu
}

\institute{Shanghai Jiao Tong University}
%


%
\maketitle              
\begin{abstract}

Emergence of artificial intelligence techniques in biomedical applications urges the researchers to pay more attention on the {\em uncertainty quantification (UQ)} in machine-assisted medical decision making. For classification tasks, prior studies on UQ are difficult to compare with each other, due to the lack of a unified quantitative evaluation metric. Considering that well-performing UQ models ought to know when the classification models act incorrectly, we design a new evaluation metric, area under {\em Confidence-Classification Characteristic} curves (AUCCC), to quantitatively evaluate the performance of the UQ models. AUCCC is threshold-free, robust to perturbation, and insensitive to the classification performance. We evaluate several UQ methods (e.g., max \textit{softmax} output) with AUCCC to validate its effectiveness. Furthermore, a simple scheme, named {\em Uncertainty Distillation (UDist)}, is developed to boost the UQ performance, where a confidence model is distilling the confidence estimated by deep ensembles. The proposed method is easy to implement; it consistently outperforms strong baselines on natural and medical image datasets in our experiments.

\end{abstract}

\section{Introduction}
``Rise of the machines'' in biomedical applications requires more research on AI interpretability, security, privacy and fairness issues. {\em Uncertainty quantification (UQ)}, as an important capability for explainable artificial intelligence (XAI), is urgently needed for safety-critical tasks, e.g., medical decision making \cite{begoli2019need}. With effective UQ models, clinicians are able to intervene in the automatic medical decision procedures when the decisions are made uncertainly. 

We focus on uncertainty quantification for classification. 
As suggested by prior study \cite{lakshminarayanan2017simple}, two aspects should be examined on UQ models: 1) {\em calibration} and 2) {\em generalization}. \textbf{Calibration} measures, how confident a UQ model is with accurate / inaccurate classification results on in-distribution inputs. However, the UQ performance should be decoupled from classification performance, which means a classifier (and the UQ model) ``may be very accurate yet miscalibrated, and vice versa'' \cite{lakshminarayanan2017simple}. As for \textbf{generalization}, it focus on whether the model is uncertain on out-of-distribution inputs. Hendrycks and Gimpel (2017) \cite{hendrycks2016baseline} develop maximum \textit{softmax} outputs in effectively identifying out-of-distribution samples. 

However, it is difficult to compare prior methods of uncertainty quantification (in classification) with each other, due to \textbf{the lack of a unified quantitative evaluation}. The evaluations are either qualitative \cite{gal2016dropout}, or highly dependent on the classification performance \cite{lakshminarayanan2017simple}. Research on out-of-distribution detection \cite{hendrycks2016baseline, liang2017enhancing} uses Receiver Operating Characteristic analysis to quantitatively evaluate the performance, yet it is not applicable on the in-distribution setting. Motivated by unifying the evaluation on in-distribution (calibration) and out-of-distribution (generalization) uncertainty quantification, we propose {\em Confidence-Classification Characteristic (CCC)} analysis, a UQ evaluation method orthogonal to classification. The area under the CCC curves (AUCCC) is a quantitative evaluation metric, which is threshold-free, robust to perturbation, and insensitive to classification performance. Our experiments on several datasets validate the effectiveness of CCC.

Another major challenge in UQ is that the ``ground truth'' of uncertainty estimates are generally not available \cite{lakshminarayanan2017simple}. Bayesian approaches in UQ, e.g., Monte Carlo dropout \cite{gal2016dropout}, thereby become prevalent. Besides, deep ensemble \cite{lakshminarayanan2017simple}, a non-Bayesian (yet probabilistic) alternative, develops a simple and scalable scheme to estimate uncertainty. These studies estimate the confidence\footnote{confidence $=$ 1 - uncertainty if uncertainty $\in [0,1]$} from the \textit{softmax} output,
Yet a key insight in this study is that classification and its confidence should be modeled separately.
Inspired from knowledge distillation \cite{hinton2015distilling}, we develop a cascade model, named {\em Uncertainty Distillation (UDist)}, to distill the uncertainty estimation from the Deep Ensembles \cite{lakshminarayanan2017simple}. Experiments empirically validate the effectiveness of this simple scheme over Deep Ensembles on UQ.

\section{Evaluating Uncertainty Quantification}
\subsection{A Discriminative View of Uncertainty Quantification}
Intuitively, the classification and uncertainty quantification is calibrated, if the classification accuracy is higher at high confidence and vice versa. Deep Ensemble \cite{lakshminarayanan2017simple} uses a curve of accuracy vs. confidence to evaluate the calibration, yet this evaluation is highly dependent on the classification model. 
We believe the evaluation of UQ is orthogonal to the classification performance. Even if classification is modeled terribly deficient, a UQ model should be regarded as {\em perfect} if it is ``confident'' to all accurate classification and ``uncertain'' to all inaccurate classification. To this regard, we propose a discriminative view of UQ, where the UQ model is discriminating between the ``accurate’‘ or ``inaccurate'' classification. 

\begin{figure}[!htb]
\includegraphics[width=\textwidth]{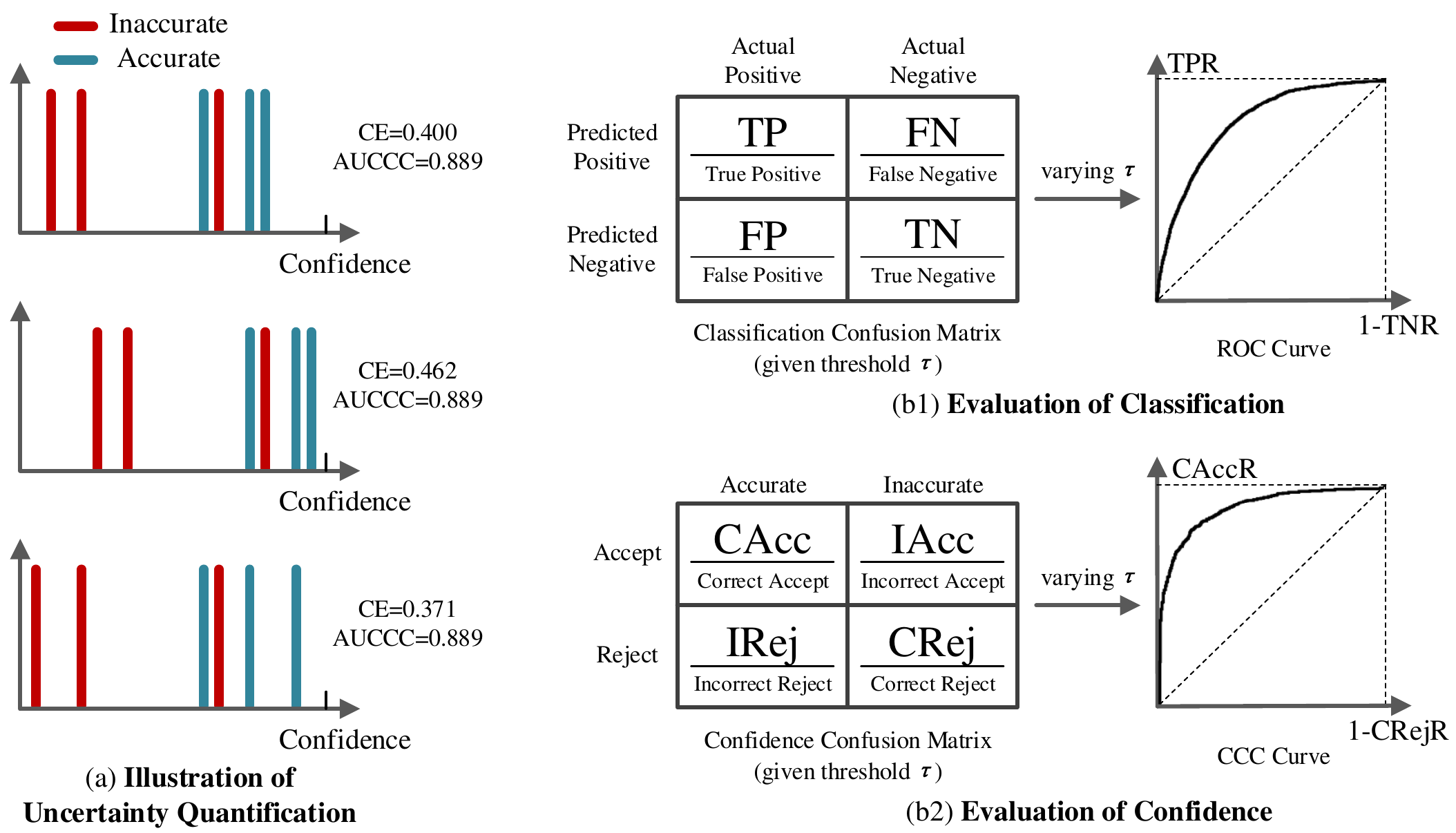}
\caption{(a) Illustration of discriminative view of uncertainty quantification. 
(b) Analogy between (b1) Receiver Operating Characteristic (ROC) and (b2) Confidence-Classification Characteristic (CCC) analysis. Similar to ROC, a unique confidence confusion matrix is constructed given a threshold $\tau$. By varying $\tau$, we get various confusion matrices. We plot the CCC curves, with $1-\mathit{CRejR}$ on the $x$-axis and $\mathit{CAccR}$ on the $Y$-axis.} 
\label{fig:2}
\end{figure}

With the help of the discriminative view, we are able to quantitatively analyze the performance of a UQ model. A UQ model is expected to separate the ``accurate'' and ``inaccurate'' classification as far as possible, by giving out various confidence score for various instances. 
As demonstrated in Figure \ref{fig:2} (a), given the results of classification and the confidence scores (assume confidence $\in [0,1]$), several metrics are able to measure this separability, e.g., cross entropy and Brier score \cite{lakshminarayanan2017simple}. We take cross entropy (CE) for illustration, by defining ``accurate'' as positive and ``inaccurate'' as negative. CE for the upper (U), middle (M) and bottom (B) cases indicates that the (M) is the worst and (B) is the best. In fact, (M) is generated by shifting all confidences in (U) to the right by a same margin, which should not be regarded as a different UQ model. Besides, the (B) is a UQ model with small perturbation to the leftmost and rightmost confidence in (U). We believe a fair evaluation for uncertainty quantification should not be sensitive to 
such small perturbation. Cross entropy and Brier score are defective measures in this sense. 

For this reason, an eligible metric should be designed to be {\em threshold-free}, {\em robust on perturbation} and {\em insensitive to classification performance}, which motivates us to consider a new evaluation metric for uncertainty quantification.

\subsection{Confidence-Classification Characteristic Analysis}





Following the discriminative view of UQ, we introduce Confidence-Classification Characteristic (CCC) analysis, which is motivated by Receiver Operating Characteristic (ROC) analysis \cite{fawcett2006introduction}.  
Formally, each instance is mapped to a class label via a classification model, and meanwhile mapped to a confidence score via a UQ model, indicating whether the classification should be accepted. Given an instance, there are \textbf{four} possible outcomes: 1) \emph{correct accept (CAcc)}, if the result is accepted on an accurate classification; 2) \emph{correct reject (CRej)}, if the result is rejected on an inaccurate classification; 3) \emph{incorrect accept (IAcc)}, if the result is accepted on an inaccurate classification; 4) \emph{incorrect reject (IRej)}, if the result is rejected on an accurate classification. Given a set of instances, a two-by-two confidence confusion matrix can be abtained, as in Figure \ref{fig:2} (b). Several metrics can be calculated based on the matrix, among which we define the two most important metrics, \emph{correct accept rate (CAccR)} and \emph{correct reject rate (CRejR)}:

\begin{equation}
\mathit{CAccR} = \frac{\mathit{CAcc}}{\mathit{CAcc} + \mathit{IRej}}\;,\qquad \mathit{CRejR} = \frac{\mathit{CRej}}{\mathit{CRej} + \mathit{IAcc}}.
\end{equation}

A two-dimensional CCC curve is plotted with \textit{CAccR} on the $Y$-axis and $1-\mathit{CRejR}$ on the $x$-axis, which denotes the trade-offs between more \textbf{acceptance} with more incorrect results and more \textbf{rejection} with less incorrect results. Similar to ROC, the CCC curve is upper if the model has better performance on confidence evaluation. To further reduce the evaluation to a single scalar value, we define the area under the CCC curve as \emph{AUCCC}. A random confidence model leads to an AUCCC of 0.5, indicating that the confidence of accurate results and inaccurate results are entirely mixed up. On the other hand, a perfect confidence model is expected to have an AUCCC of 1.0, representing that each accurate result have a confidence score higher than the inaccurate results. In more general circumstances, the CCC curve is convex with an AUCCC score $\in [0.5,1]$. The CCC analysis elegantly unifies the in-distribution and out-of-distribution uncertainty quantification, by assigning all classification results on the out-of-distribution instances as ``inaccurate classification''. It is worth noting that, even for binary classification, the CCC and ROC measure different aspects. ROC evaluates how accurate the predictions are, given various classification thresholds; CCC focuses on how accurate the \emph{accepted} predictions are, with regard to various \emph{confidence} thresholds, given a \textbf{fixed} classification threshold (e.g., 0.5).

Essentially, CCC is a special instance of ROC analysis (refer to Figure \ref{fig:2} (b) for illustration), where confidence serves as the score function, and ``accurate'' / ``inaccurate'' classification is the positive / negative class. Hence CCC inherits several advantages for ROC analysis. First, CCC is threshold-free, by taking all possible thresholds to generate the CCC curve. Second, CCC is robust on perturbation; subtle perturbation on confidence scores does not affect the CCC curve, without violation of the order of all scores. Third, considering that ROC is insensitive to class imbalance \cite{fawcett2006introduction}, CCC is insensitive to classification performance, which decouples the evaluation of UQ from classification models. 


\section{Boosting UQ Performance via Uncertainty Distillation}

\begin{figure}[!htb]
\includegraphics[width=\textwidth]{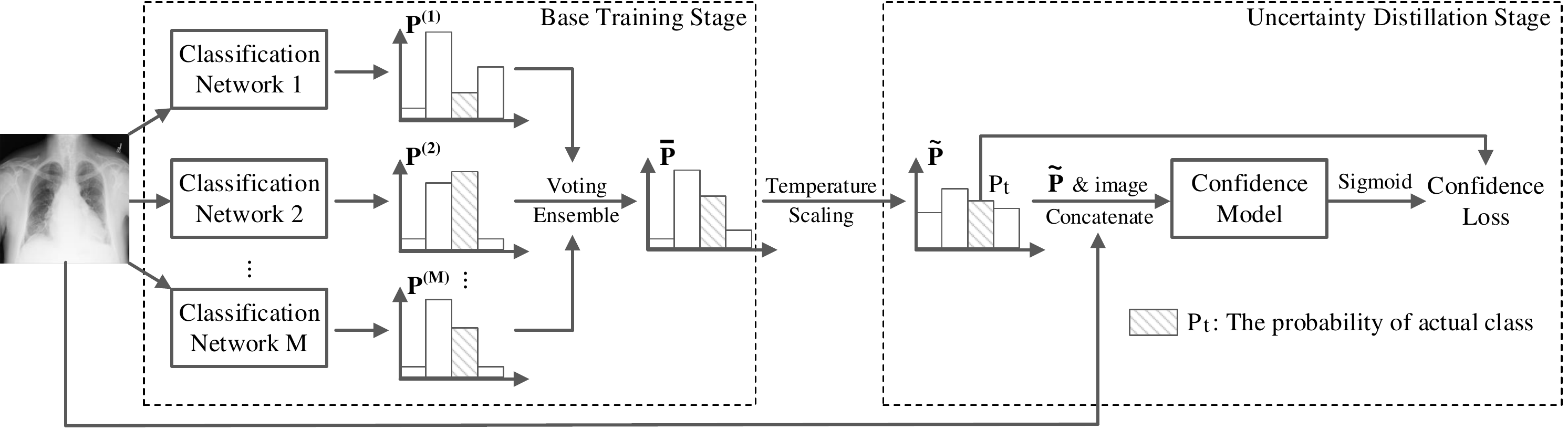}
  \caption{The cascade for Uncertainty Distillation. In base training stage, $M$ classification models are trained independently. The $M$ outputs are then averaged and temperature scaled, before concatenated with the input image to be fed into the confidence model in uncertainty distillation stage. The softened ensemble output of the actual class ($P_t$) serves as the ``ground truth'' in the confidence loss.} 
\label{fig:1}
\end{figure}


Prior research couples the modeling of classification and confidence in single models. Even though the confidence extracted from classification output is theoretically calibrated\footnote{It is proven \cite{gneiting2007strictly} that the outputs optimized with a proper scoring rule, e.g., cross entropy loss, are calibrated with ``confidence''.} for classification \cite{gneiting2007strictly, lakshminarayanan2017simple}, there is no trivial method to extract the output of the \textbf{actual class} in the inference stage, due to the lack of ground truth. Max \textit{softmax} output is proposed \cite{hendrycks2016baseline} as an alternative, it is however over-confident at inaccurate classification. Inspired from knowledge distillation \cite{hinton2015distilling}, where a student model is better optimized with teachers' outputs as ground truth, we propose to distill the uncertainty estimated by a deep ensemble \cite{lakshminarayanan2017simple}. This simple scheme, named Uncertainty Distillation (UDist), is effective in boosting the UQ performance. 

As depicted in Figure \ref{fig:1}, in base training stage, $M$ classification models are trained on the datasets independently, from which \textit{softmax} probability distributions $P^{(i)}\,(i=1,2,...,M)$ are obtained on each instance. Due to over-fitting, the probabilities on the training set are generally higher than that on test set, which we call the {\em over-confidence} issue. To alleviate the over-confidence issue of single model, an ensemble probability $\overline{P}$ is averaged for better calibration \cite{lakshminarayanan2017simple}. We then apply temperature scaling to further soften the over-confident output:
\begin{equation}
    \widetilde{P} = \mathit{softmax}(\log(\overline{P}) / T).
\end{equation}


We denote $P_t$ as the output of actual class from the softened ensemble output $\widetilde{P}$, which represents the (softened) confidence estimated by the deep ensemble. $P_t$ is generally unavailable in the inference stage, due to the lack of ground truth. To this end, we optimize the confidence score $s$ towards the confidence estimate $P_t$ by a cross entropy loss, also named as the {\em confidence loss} in our study,
\begin{equation}
    loss_{conf}(s, P_t) = P_t \times \log(s) + (1-P_t) \times \log(1-s).
\end{equation}

In the uncertainty distillation stage, a cascade model is designed as the confidence model, whose inputs are concatenation of the input images and softened probability $\widetilde{P}$ from deep ensemble. Only the input images are insufficient, since the confidence is jointly dependent on the input and the output. Theoretically, the UDist cascade is able to identify the inaccurate classification, with the confidence loss and not over-confident $P_t$ as ground truth.


\section{Experiments}

We validate the effectiveness of the proposed methods on a widely-used natural image dataset CIFAR-10, and a multi-label medical image dataset ChestX-ray14 \cite{wang2017chestx}. AUCCC is valid on evaluating several baselines, and the proposed UDist cascade consistently outperforms the baselines. Finally, we address evaluation on identifying in-distribution (\emph{id}) and out-of-distribution (\emph{ood}) samples.
\subsection{CIFAR-10}
\subsubsection{Experiment Settings}
CIFAR-10 consists of $60,000$ colored natural images of $10$ classes, among which $50,000$ for training and $10,000$ for testing. A standard data normalization and augmentation scheme that is widely used for this dataset is adopted. We set up several baselines: 1) VGG \cite{simonyan2014very}; 2) DenseNet \cite{huang2017densely}; 3) deep ensemble on \textit{four} DenseNets ($T=1$); 4) deep ensembles with temperature scaling ($T=3$), proposed by Liang and Li \cite{liang2017enhancing}. For UDist, two DenseNets are adopted respectively for classification model and confidence model, trained with $T=8$ and tested with $T=3$. Each DenseNet shares the same structure: dense layers are repeated $[16,16,16]$ times before down-sampling, with a growth rate of $12$, resulting in a model size of $3.3Mb$. VGG and DenseNet also share about the same number of parameters. Since classification performance is not our issue, we employ small networks due to computation constraints. 
 We use an Adam optimizer \cite{kingma2014adam} for all training with a batch size of $256$,  with an initial learning rate $1\times 10^{-3}$. All classification models are trained for $300$ epochs, while the confidence model is trained for only $20$ epochs since it converges extremely fast. We decay the learning rate by $10^{-1}$ at $1/2$ and $3/4$ milestones of all epochs. Accuracy (not including UDist) and AUCCC on test set are reported. 
\begin{table}[!htb]
        \centering
        \caption{Classification and UQ performance on CIFAR-10 dataset.}
    	\begin{tabular}{cccccc}
	    	\toprule
	    	\multirow{2}{*}{Metrics}   & \multirow{2}{*}{\;\;VGG\;\;} & \multirow{2}{*}{\;\;DenseNet\;\;} & Deep Ensemble\;\;& Deep Ensemble\;\;  & \multirow{2}{*}{\;\;UDist} \\ 
	    	&&& ($M=4, T=1$) & ($M=4, T=3$) &\\
		    \midrule
		    Classification Acc\;\; & 90.80 & 91.61 & 92.99 & 93.00 & -\\
            Confidence AUCCC\;\; & 90.48 & 91.21 & 92.75 & 93.03 & \textbf{93.56} \\
		    \bottomrule
    	\end{tabular}
        \label{CIFAR-10-performance}
\end{table}
\vspace{-10px}
\subsubsection{Results}
As illustrated in Table \ref{CIFAR-10-performance}, the improvement on classification also leads to the improvement on UQ performance. VGG has the worst performance due to its simple structure. Deep ensemble based on DenseNets can improve both AUROC and AUCCC by a large margin. Significantly, UDist can further improve UQ performance over ensemble by temperature scaling.
\subsection{ChestX-ray14}
\subsubsection{Experiment Settings}
In this experiment, we use the NIH ChestX-ray14 dataset \cite{wang2017chestx}, which consists of $112,120$ X-ray images of $30,805$ patients. Each patient is labeled with $14$ deceases. The multi-label confidence loss averaged over confidence loss of all classes. The models' configuration and training strategies are the same as that of the CIFAR-10 experiment, resulting in a model size of $3.3Mb$.
\begin{table}[!htb]
\vspace{-10px}
        \centering
        \caption{Classification and UQ performance on ChextX-ray14 dataset.}
    	\begin{tabular}{cccccc}
	    	\toprule
	    	\multirow{2}{*}{Metrics}   & \multirow{2}{*}{\;\;VGG\;\;} & \multirow{2}{*}{\;\;DenseNet\;\;} & Deep Ensemble\;\;& Deep Ensemble\;\;  & \multirow{2}{*}{\;\;UDist} \\ 
	    	&&& ($M=4, T=1$) &($M=4, T=10$)&\\
		    \midrule
		    Classification AUC & 81.50 & 82.37 & 82.86 & 82.86 & -\\
            Confidence AUCCC & 81.34 & 82.07 & 82.56 & 82.56 & \textbf{82.73} \\
		    \bottomrule
    	\end{tabular}
        \label{ChextX-ray14-performance}
\vspace{-10px}
\end{table}
\subsubsection{Results}
UDist outperforms all baselines in uncertainty quantification. Note that deep ensemble provides $0.5\%$ advancement over DenseNet on this dataset, while our model achieves even more significant improvement by a margin of $0.7\%$. 
\subsection{On Out-of-Distribution Samples}
\subsubsection{Experiment Settings}
Deep models are expected to provide low confidence, when the test data is distinctive from the training data \cite{hendrycks2016baseline}. With better UQ performance, the model distinguishes better out-of-distribution (\emph{ood}) data from in-distribution (\emph{id}) data. We conducted experiments on two sets of datasets. First, we evaluate models trained on CIFAR-10 with CIFAR-10 ({\em id}) vs. the Street View House Numbers (SVHN) \cite{netzer2011reading}({\em ood}). Second, we evaluate models trained on ChestX-ray14 with ChestX-ray14 ({\em id}) vs. OCT2017\footnote{A dataset with optical coherence tomography (OCT) images of the retina.} dataset \cite{kermany2018identifying}({\em ood}).
Both experiments randomly select the same numbers of \emph{ood} samples as \emph{id}. Deep ensemble with temperature scaling proposed by Liang and Li \cite{liang2017enhancing} is a strong baseline compared with UDist ($T=10$). We reports two kinds of metrics. The first one is our AUCCC metric, where predictions on \emph{ood} samples are labeled with \emph{inaccurate}. The CCC unifies the \emph{id} and \emph{ood} uncertainty quanfication. Second, to adapt the same experiment settings from prior study \cite{hendrycks2016baseline}, we label predictions on \emph{id} samples with \emph{accurate} and those on \emph{ood} with \emph{inaccurate}. We apply ROC to analyze how separate the \emph{id} and \emph{ood} samples are, resulting in a metric named I/O AUROC. Higher value in these two metrics indicates better out-of-distribution (generalization) UQ performance. 
\begin{table}[!htb]
        \centering
        \caption{UQ performance on identifying out-of-distribution samples.}
    	\begin{tabular}{cccccc}
		  	\toprule
		  	  & \multicolumn{2}{c}{CIFAR-10 vs. SVHN} & \;\;& \multicolumn{2}{c}{ChestX-ray14 vs. OCT2017}  \\
	    	Metrics\;\;\;\;   &  Deep Ensemble & \;\;\;\;UDist\;\;\;\; & & Deep Ensemble & \;\;\;\;UDist\;\;\;\;\\ 
		    \midrule
		   I/O AUROC\;\;\;\; & 97.76 & \textbf{98.18} & & 68.09 & \textbf{69.42} \\
            AUCCC\;\;\;\; & 98.04 & \textbf{98.25} & & 70.61 & \textbf{71.77}\\
		    \bottomrule
    	\end{tabular}
        \label{Ood-performance}
\vspace{-10px}
\end{table}
\subsubsection{Results}
As depicted in Table \ref{Ood-performance}, UDist is effective to distinguish \emph{ood} instances from \emph{id} instances in all kinds of settings, indicating its excellent performance in uncertainty quantification. Besides, we figure out two more important findings: 1) AUCCC metric is generally higher than I/O AUROC. We argue that it is because AUCCC considers accurate and inaccurate results on \textit{id} samples separately, which is more eligible than I/O AUROC. 2) UDist achieves more significant improvement using I/O AUROC metric, i.e., when we consider \textit{ood} and \textit{id} generalization alone. We argue that these improvement comes from UDist's advantage to distinguish \textit{ood} samples from \textit{id} samples.
\section{Discussions}
\subsubsection{Over-confidence issue.} A major challenge for the uncertainty distillation is the over-confidence issue on the training set. In our study, we apply ensemble and temperature scaling techniques to alleviate this problem. However, deep neural networks still tend to overfit the training set; on CIFAR-10, a base model achieves an AUCCC over 0.99, much more higher than that on the test set. Our Uncertainty Distillation cascade demonstrates promising results even trained with over-confident uncertainty estimates. In future study, independent datasets on uncertainty estimates will be involved to solve the over-confidence issue.
\vspace{-10px}
\subsubsection{On the cascade inputs.} In our current cascade model, classification information is directly concatenated with the image input. It decays after layers of convolution and normalization, and becomes hard to associate with the final outputs, especially for multi-label classification (e.g., ChestX-ray14). Advanced techniques in conditional generation may benefit this problem.

\section{Conclusion}
In this study, we propose a new evaluation method, Confidence-Classification Characteristic analysis, which unifies the in-distribution (calibration) and out-of-distribution (generalization) uncertainty quantification. Moreover, a cascade model, named Uncertainty Distillation, is proposed to boost the performance of uncertainty quantification over strong baselines. In future study, we are solving the over-confidence issue, and simplifying the cascade structure.
\bibliographystyle{splncs04}
\bibliography{ref}

\begin{thebibliography}{10}
\providecommand{\url}[1]{\texttt{#1}}
\providecommand{\urlprefix}{URL }
\providecommand{\doi}[1]{https://doi.org/#1}

\bibitem{begoli2019need}
Begoli, E., Bhattacharya, T., Kusnezov, D.: The need for uncertainty
  quantification in machine-assisted medical decision making. Nat. Mach.
  Intell.  \textbf{1}(1), ~20 (2019)

\bibitem{fawcett2006introduction}
Fawcett, T.: An introduction to roc analysis. Pat. Rec. Lett.  \textbf{27}(8),
  861--874 (2006)

\bibitem{gal2016dropout}
Gal, Y., Ghahramani, Z.: Dropout as a bayesian approximation: Representing
  model uncertainty in deep learning. In: ICML. pp. 1050--1059 (2016)

\bibitem{gneiting2007strictly}
Gneiting, T., Raftery, A.E.: Strictly proper scoring rules, prediction, and
  estimation. Journal of the American Statistical Association
  \textbf{102}(477),  359--378 (2007)

\bibitem{hendrycks2016baseline}
Hendrycks, D., Gimpel, K.: A baseline for detecting misclassified and
  out-of-distribution examples in neural networks. In: ICLR (2017)

\bibitem{hinton2015distilling}
Hinton, G., Vinyals, O., Dean, J.: Distilling the knowledge in a neural
  network. In: NIPS Deep Learning Workshop (2014)

\bibitem{huang2017densely}
Huang, G., Liu, Z., Van Der~Maaten, L., Weinberger, K.Q.: Densely connected
  convolutional networks. In: CVPR. pp. 4700--4708 (2017)

\bibitem{kermany2018identifying}
Kermany, D.S., Goldbaum, M., Cai, W., et~al.: Identifying medical diagnoses and
  treatable diseases by image-based deep learning. Cell  \textbf{172}(5),
  1122--1131 (2018)

\bibitem{kingma2014adam}
Kingma, D.P., Ba, J.: Adam: A method for stochastic optimization. In: ICLR
  (2014)

\bibitem{lakshminarayanan2017simple}
Lakshminarayanan, B., Pritzel, A., Blundell, C.: Simple and scalable predictive
  uncertainty estimation using deep ensembles. In: NIPS. pp. 6402--6413 (2017)

\bibitem{liang2017enhancing}
Liang, S., Li, Y., Srikant, R.: Enhancing the reliability of
  out-of-distribution image detection in neural networks. In: ICLR (2018)

\bibitem{netzer2011reading}
Netzer, Y., Wang, T., Coates, A., Bissacco, A., Wu, B., Ng, A.Y.: Reading
  digits in natural images with unsupervised feature learning  (2011)

\bibitem{simonyan2014very}
Simonyan, K., Zisserman, A.: Very deep convolutional networks for large-scale
  image recognition. In: CVPR (2014)

\bibitem{wang2017chestx}
Wang, X., Peng, Y., Lu, L., et~al.: Chestx-ray8: Hospital-scale chest x-ray
  database and benchmarks on weakly-supervised classification and localization
  of common thorax diseases. In: CVPR. pp. 2097--2106 (2017)

\end{thebibliography}
%




\end{document}